\documentclass{article}

\usepackage{arxiv}

\usepackage[utf8]{inputenc} 
\usepackage[T1]{fontenc}    
\usepackage{hyperref}       
\usepackage{url}            
\usepackage{booktabs}       
\usepackage{amsfonts}       
\usepackage{nicefrac}       
\usepackage{microtype}      
\usepackage{cleveref}       
\usepackage{lipsum}         
\usepackage{graphicx}
\usepackage{natbib}
\usepackage{doi}

\usepackage{url,hyperref,lineno,microtype,subcaption}
\usepackage[onehalfspacing]{setspace}
\usepackage{makecell}
\usepackage{amstext}

\def\etal{\emph{et al. }}

\title{Selective Masking based Self-Supervised Learning for Image Semantic Segmentation}

\date{}

\author{ {Yuemin Wang} \\
	Department of Computer Science\\
	University of Sasktachewan\\
	Saskatoon, SK, Canada \\
	\texttt{yuw422@usask.ca} \\
	\And
	{Ian Stavness} \\
	Department of Computer Science\\
	University of Sasktachewan\\
	Saskatoon, SK, Canada \\
	\texttt{ian.stavness@usask.ca} \\
}

\graphicspath{ {images/} }

\begin{document}

\maketitle

\begin{abstract}
\textbf{Introduction}: This paper proposes a novel self-supervised learning method for semantic segmentation using selective masking image reconstruction as the pretraining task. Our proposed method replaces the random masking augmentation used in most masked image modelling pretraining methods. 

\textbf{Methods}: The proposed selective masking method selectively masks image patches with the highest reconstruction loss by breaking the image reconstruction pretraining into iterative steps to leverage the trained model’s knowledge. 

\textbf{Results}: We show on two general datasets (Pascal VOC and Cityscapes) and two weed segmentation datasets (Nassar 2020 and Sugarbeets 2016) that our proposed selective masking method outperforms the traditional random masking method and supervised ImageNet pretraining on downstream segmentation accuracy by 2.9\% for general datasets and 2.5\% for weed segmentation datasets. Furthermore, we found that our selective masking method significantly improves accuracy for the lowest-performing classes. Lastly, we show that using the same pretraining and downstream dataset yields the best result for low-budget self-supervised pretraining.

\textbf{Discussion}: Our proposed Selective Masking Image Reconstruction method provides an effective and practical solution to improve end-to-end semantic segmentation workflows, especially for scenarios that require limited model capacity to meet inference speed and computational resource requirements. 

\end{abstract}


\section{Introduction}
\label{sec:intro}

Self-supervised learning (SSL) has played a key role in the recent success of large language models (LLM) \citep{bengio2013representation, brown2020language, radford2018improving, radford2019language, devlin2018bert} and large vision models (LVM) like Vision Transformers (ViT) \citep{dosovitskiy2020image} and multimodal models like CLIP \citep{radford2021learning}. Masked image modeling (MIM) is a proven method for self-supervised vision model pretraining where the model is pretrained to reconstruct randomly masked image tokens \citep{he2022masked, bao2021beit} and demonstrated state-of-the-art downstream performance on several vision tasks. Unfortunately, the performance gains obtained by large-scale self-supervised pretraining have diminishing returns as model parameters and pretraining dataset sizes continue to scale up \citep{udandarao2024no}. This motivates the need for novel masking procedures to extend the utility of MIM for pretraining.

The tremendous pretraining cost of large vision models is not justified for many computer vision applications, where generalization and accuracy must be balanced by computational efficiency and other metrics due to application requirements and hardware limitations. Many computer vision (CV) tasks, especially detection/classification tasks, value data and computational efficiency more than generalizability or accuracy. Tasks like autonomous driving are hard real-time systems where the system needs to respond within a time limit, resulting in these systems often running locally on less powerful embedded hardware to avoid instability in the wireless internet and server latency. The lack of processing power and the need for fast inference makes lightweight convolutional neural networks (CNN) like MobileNet \citep{howard2017mobilenets} and EfficientNet \citep{tan2019efficientnet} still compelling choices for many applications. CNNs are also known to require considerably fewer data samples to train compared to ViTs \citep{zhao2021battle}, where they remain competitive in areas suffering from data scarcity like medical imaging
\citep{ronneberger2015u}.

For semantic segmentation tasks, fully-supervised pretrained models, e.g. using ImageNet \citep{deng2009imagenet} pretrained backbones, are generally considered the gold standard \citep{he2019rethinking} for most downstream tasks, but are known to suffer from domain shift \citep{kim2022broad}. Attempts to tackle domain shift led to data-efficient SSL approaches where models can be effectively pretrained on relatively sizeable in-domain unlabelled datasets \citep{gao2022convmae, xie2024rethinking}. Utilizing in-domain unlabeled images has a practical benefit, because when creating image datasets, it is common to collect considerably more images than one can afford to label. This is especially true for semantic segmentation datasets, where pixel-level class labels are time-consuming and tedious to annotate, making the cost to acquire images much lower than the cost to annotate. For example, the Pascal VOC \textsl{2012} dataset \citep{everingham2015pascal} contains 17,125 images in total, with only 2,913 images labeled for semantic segmentation. Learning representations from unlabelled images could result in better downstream performance or lower labelling costs.

In this paper, we study generative SSL methods for data-efficient semantic segmentation models, more specifically, masked image modelling (MIM), which took inspiration from masked token prediction \citep{devlin2018bert} used in LLM pretraining. Naturally, most proposed MIM methods \citep{he2022masked, bao2021beit} are designed for transformer-based models, while generative SSL methods for CNNs, which are very relevant in efficient industrial applications, are largely overlooked.

\begin{figure}[h!]
\centering
\begin{minipage}{\textwidth}
	\includegraphics[height=1.45cm]{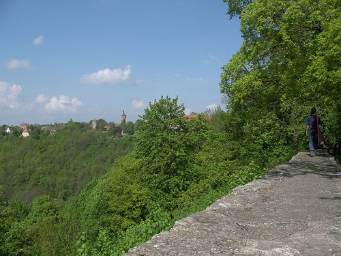}%
	\includegraphics[height=1.45cm]{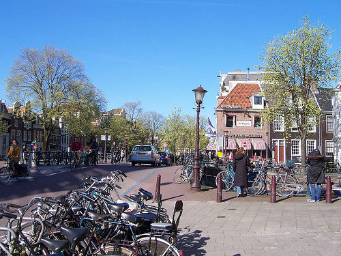}\hfill
	\includegraphics[height=1.45cm]{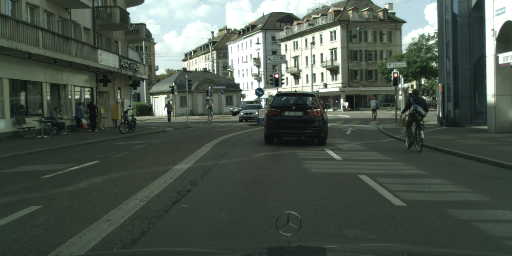}%
	\includegraphics[height=1.45cm]{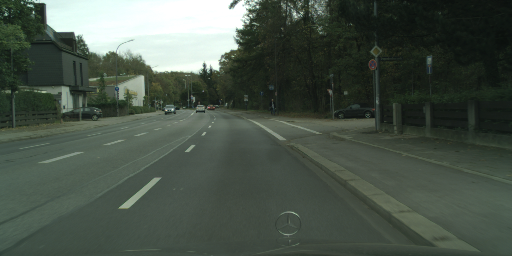}\hfill
	\includegraphics[height=1.45cm]{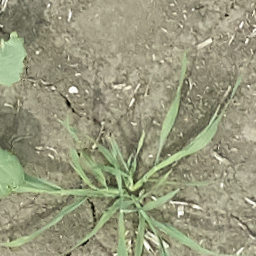}%
	\includegraphics[height=1.45cm]{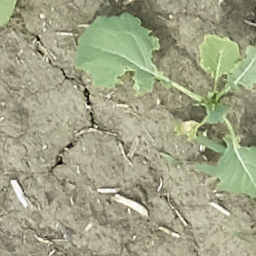}\hfill
	\includegraphics[height=1.45cm]{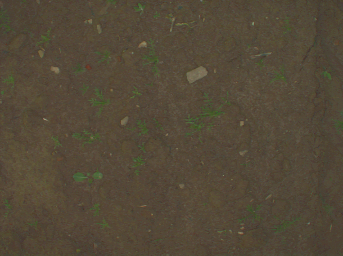}%
	\includegraphics[height=1.45cm]{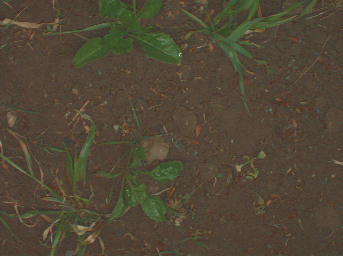}\\
	\centering
	original image \\
	\vspace{4pt}
	\includegraphics[height=1.45cm]{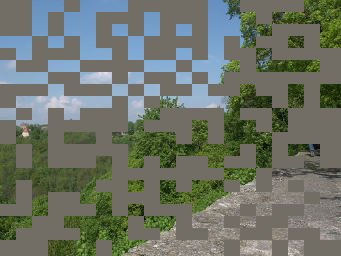}%
	\includegraphics[height=1.45cm]{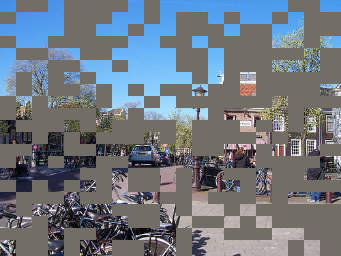}\hfill
	\includegraphics[height=1.45cm]{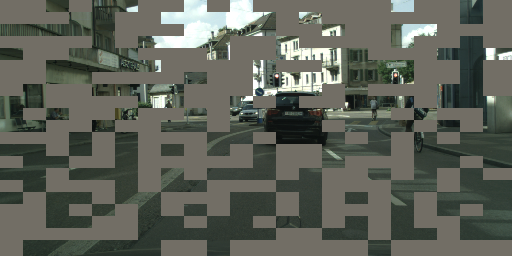}%
	\includegraphics[height=1.45cm]{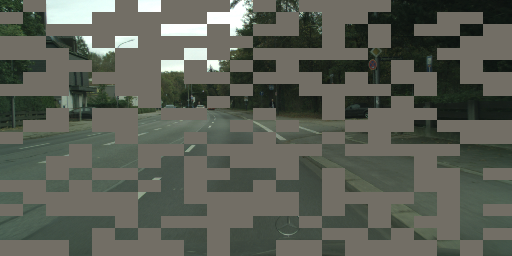}\hfill
	\includegraphics[height=1.45cm]{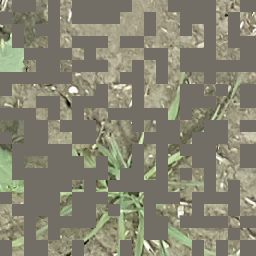}%
	\includegraphics[height=1.45cm]{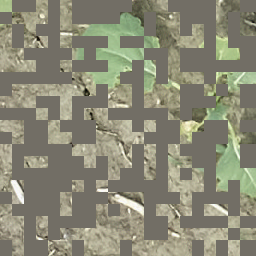}\hfill
	\includegraphics[height=1.45cm]{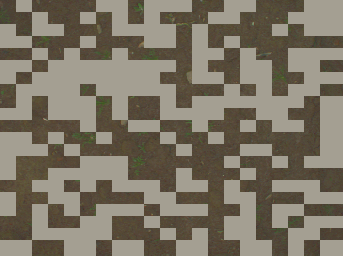}%
	\includegraphics[height=1.45cm]{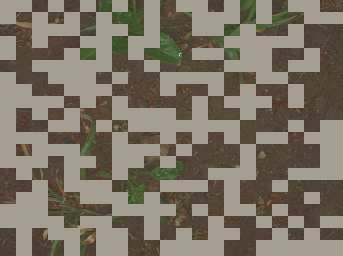}\\
	\centering	
	\vspace{-4pt}
	random masking \\
	\vspace{4pt}
	\includegraphics[height=1.45cm]{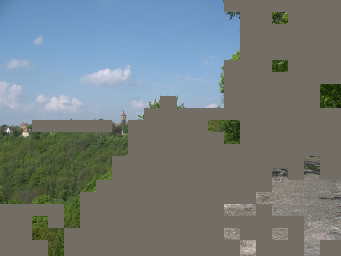}%
	\includegraphics[height=1.45cm]{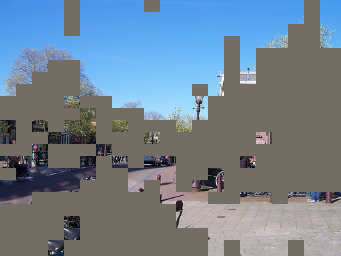}\hfill
	\includegraphics[height=1.45cm]{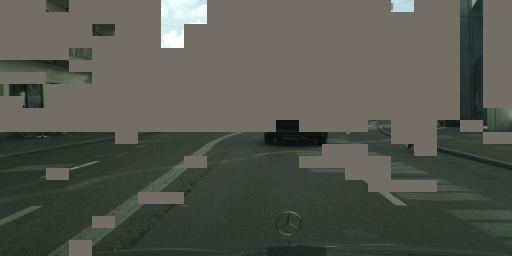}%
	\includegraphics[height=1.45cm]{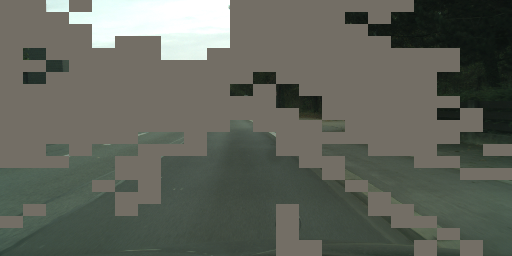}\hfill
	\includegraphics[height=1.45cm]{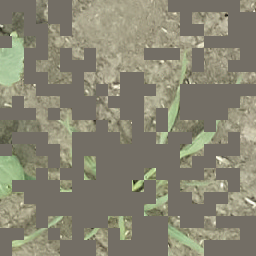}%
	\includegraphics[height=1.45cm]{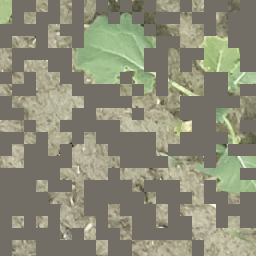}\hfill
	\includegraphics[height=1.45cm]{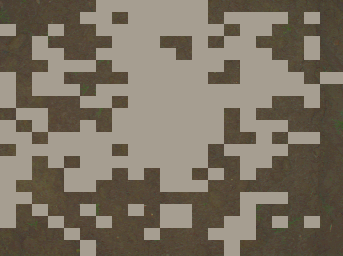}%
	\includegraphics[height=1.45cm]{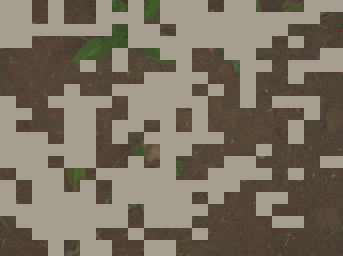}\\
	\centering
	\vspace{-4pt}
	selective masking \\
\end{minipage}

\caption[Examples comparing randomly and selectively masked images.]{Examples comparing randomly and selectively masked images. Left-to-right: two examples each for Pascal VOC, Cityscapes, Nassar2020, and Sugarbeets 2016. The brightness of Sugarbeets images is increased to improve visibility.  We can observe selective masking following the edges of leaves and roads in examples in the second column. Other examples show that selective masks are generally concentrated on non-background objects.}
\label{fig:comp_mask}
\end{figure}

Self-supervised MIM methods typically learn useful spatial, color and textural representations from reconstructing randomly masked image tokens/patches (Figure \ref{fig:comp_mask}, second row). The learned representations are, in turn, highly relevant to the success of downstream tasks. Large-scale MIM pretraining learns effectively from abundant randomly masked images augmented from their large datasets. This random masking pretraining process is typically data-hungry and intuitively contradicts how humans efficiently learn: After reading a textbook, we typically only review difficult sections that we do not understand instead of randomly reviewing sections until the entire textbook is covered. We show in this paper that random masking for low-budget MIM pretraining is not effective due to its low learning efficiency and leveraging model knowledge to selectively mask images (Figure \ref{fig:comp_mask}, row three) can significantly improve the downstream performance for low-budget MIM pretraining.

We propose Selective Masking Image Reconstruction, a novel CNN-compatible generative SSL method that selectively masks pretraining images using the model’s learnt knowledge iteratively. We show that our selective masking pretraining yields better downstream performance than both random masking and ImageNet pretraining on four distinct datasets: Pascal VOC \citep{everingham2015pascal}, Cityscapes \citep{cordts2016cityscapes}, Nassar 2020 \citep{wang2023weed}, and Sugarbeets 2016 \citep{chebrolu2017ijrr}. Our method shows clear advantages in both mean intersection-over-union (mIoU) over all classes and intersection-over-union (IoU) for the lowest-performing classes. Lastly, we show that pretraining on the same dataset as the downstream task (self-pretraining) leads to better performance in this low-budget pretraining scenario than pretraining on another dataset (cross-pretraining).

Our code is publicly available at \url{https://github.com/yuw422/Selective_Masking_Image_Reconstruction.git}.
The contributions of our paper include:

\let\labelitemi\labelitemii

\begin{enumerate}
\item We propose a novel Selective Masking Image Reconstruction SSL method that uses model knowledge to improve masking effectiveness.
\item We show that our proposed selective masking SSL method outperforms random masking and supervised ImageNet pretraining on four datasets.
\item We show our proposed selective masking SSL method boosts accuracy for lowest-performing classes.
\item We show that self-pretraining yields better performance in a low-budget pretraining scenario.

\end{enumerate}

\section{Methods}

\begin{figure}[h!]
\centerline{\includegraphics[width=1.0\linewidth]{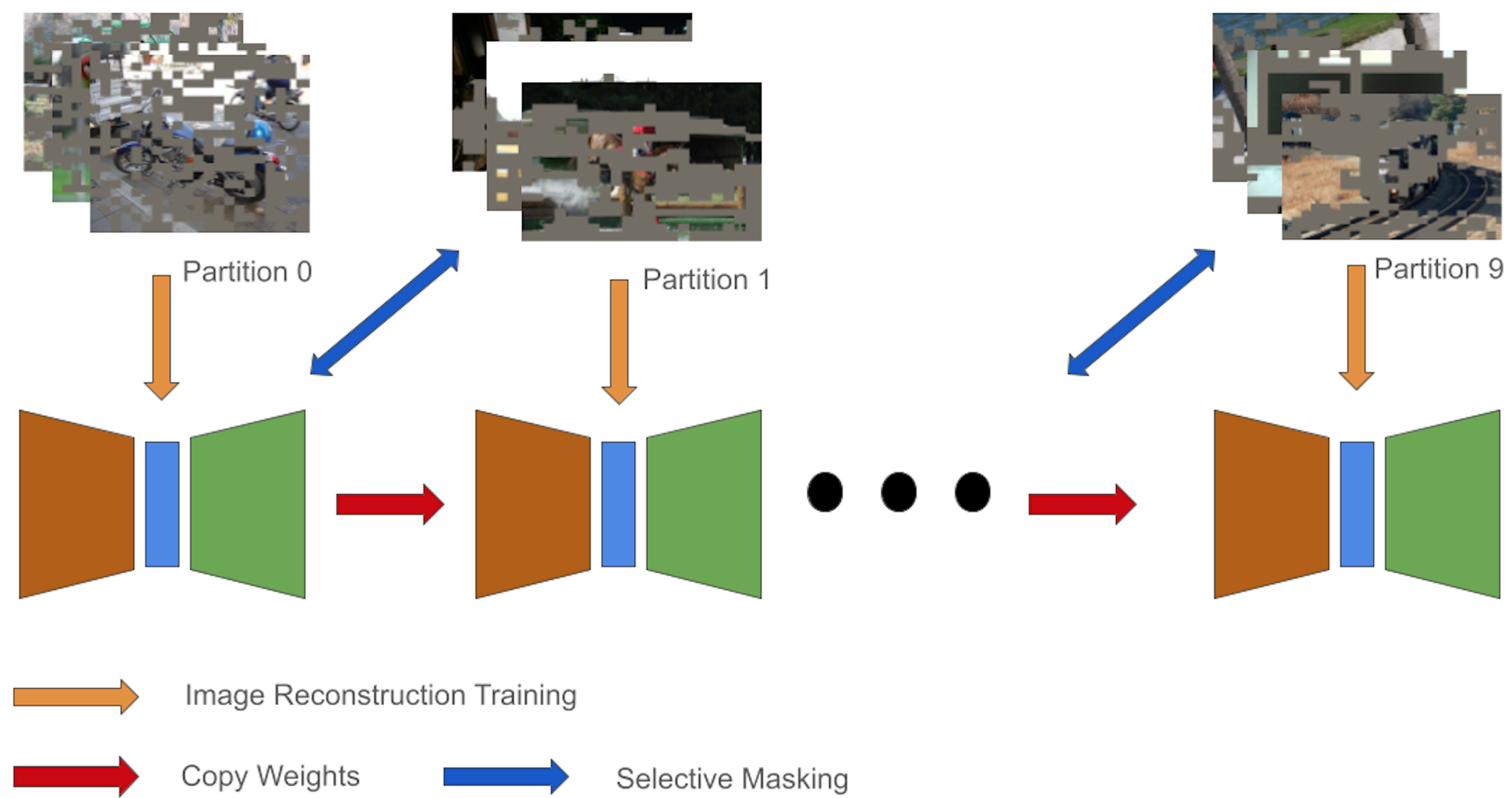}}
\caption[Our proposed Selective Masking Image Reconstruction SSL method.]{Our proposed Selective Masking Image Reconstruction SSL method iteratively trains on ten equally divided partitions of the pretraining dataset. The model is initialized by training on randomly masked partition 0. We then selectively mask partition $i$ using the model trained on partition $i-1$. Model weights from partition $i-1$ training are used to initialize the model for partition $i$, which is then trained on selectively masked partition $i$. We repeat this selective masking and training process until all partitions are used.}
\label{fig:ssl}
\end{figure}

Our proposed selective masking pretraining is different from traditional random masking methods where instead of training the model on the entire dataset, we split the dataset into ten equal partitions and iteratively train our model on each partition as shown in Figure \ref{fig:ssl}. The model trained on the previous partition is used to reconstruct randomly masked images in the current partition. We then selectively mask the patches with the highest loss values for each image in the current partition and then continue to train the model on this selectively masked partition. The pretraining is completed when we exhaust all partitions and the pretraining weights will be used to initialize the downstream training.

The pretraining process consists of one initialization step and nine selective masking steps. The initialization step trains our model to reconstruct randomly masked partition 0 images. We use the same random masking process in the initialization step, the selective masking steps and our random masking baselines. Each image is equally split into 512 patches, and 50\% of the patches are randomly masked as shown in Figure \ref{fig:comp_mask}. He \etal \citep{he2022masked} suggested a high masking percentage of 75\% but our preliminary experiment shows 75\% random masking is too high for our U-Net model resulting in poor reconstruction results and downstream performance. Figure \ref{fig:comp_mask} shows our selective masks are more condensed compared to random sampling but are not as concentrated as the block-wise masking discussed by He \etal \citep{he2022masked}, which the author recommended 50\% masking. We split images into 512 patches so that each patch is larger than 11$\times$11 pixels, which is the window size we used to calculate our multiscale-structural similarity (MS-SSIM) \citep{wang2003multiscale} loss.

\subsection{Selective Masking}
The intuition behind our selective masking follows how a human learns: only reviewing difficult sections in a textbook that we did not understand is often more efficient than reading a random chapter every time. Similarly, a model learns more from attempting to reconstruct an effective sample like a cat with complex shapes, colors and textures than the blue sky background. Random masking pretraining works on large datasets because the random sampling of millions of images can still yield enough effective samples. When pretraining on smaller datasets, it is more efficient to selectively mask the more effective portion of an image that our model struggles to reconstruct.

\begin{figure}[h!]
\centerline{\includegraphics[width=1.0\textwidth]{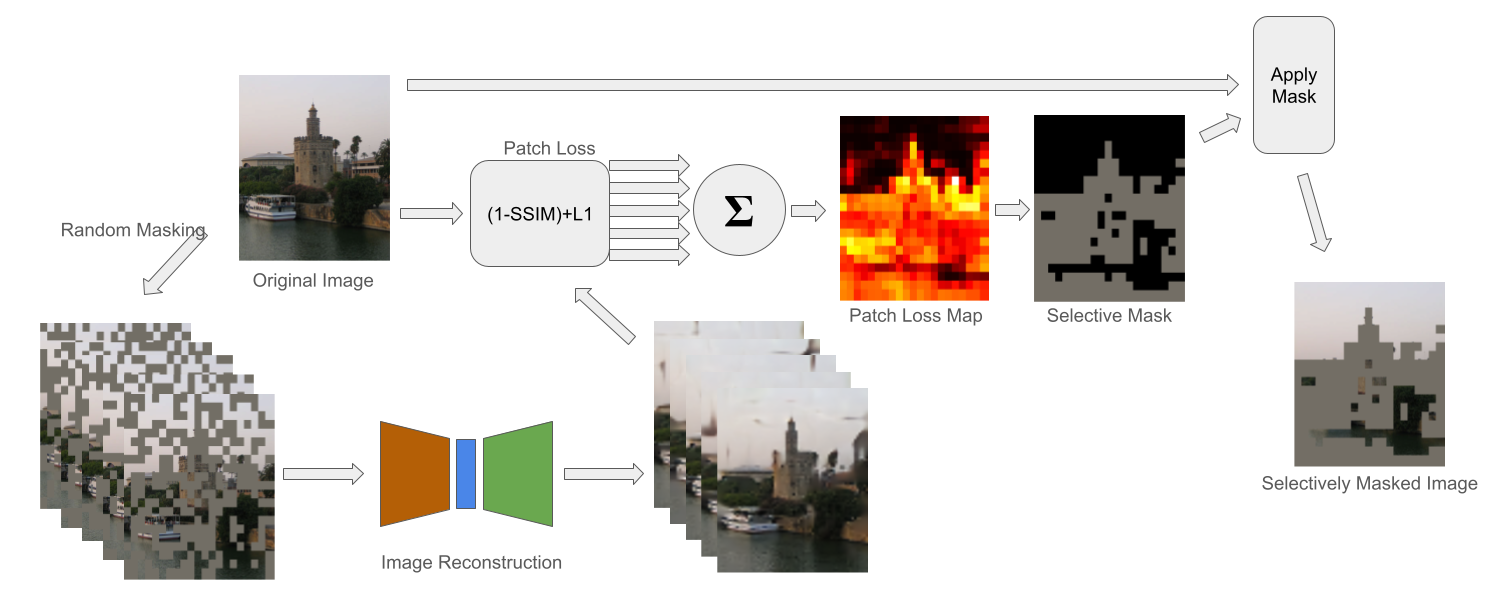}}
\caption[Our proposed selective masking process.]{Our proposed selective masking process starts by generating five 50\% randomly masked samples, which are then reconstructed by the trained U-Net model. A patch loss map is calculated for each reconstructed sample, and then the sample patch loss maps are aggregated and thresholded (top 50\%) to generate a selective mask. The selective mask is applied to the original image, resulting in the desired selectively masked image.}
\label{fig:sel_mask}
\end{figure}

Our selective masking process starts with generating five samples of an image, each with 50\% randomly masked patches as shown in Figure \ref{fig:sel_mask}. Since the 50\% unmasked patches will always have lower reconstruction loss, we create five random samples to ensure each patch in the image is masked. By doing so, the average patch loss of each patch is the desired masked reconstruction loss instead of the general model noise that makes reconstructed images blurry. Then we use our model trained on the last partition to reconstruct each sample and calculate the per-patch loss with respect to the original unmasked image. We use an SSIM \citep{wang2004image} and L1 combo loss with 3$\times$3-pixel windows to calculate the patch loss:
\begin{equation}
L_{i}(n) = (1-SSIM(M^{n}_{i}, I^{n}) )+ L1(M^{n}_{i}, I^{n})
\end{equation}
We denote the masked sample image as \(M\) with \(n \in [0,512)\) patches and \(i \in [0,5)\) samples, and the original unmasked image as \(I\). The patch loss \(L(n)\) is aggregated across samples as \(\sum_{i}L_{i}(n)\). This combo loss is similar to the loss we use to train our image reconstruction model except the training loss uses MS-SSIM with a window size of 11$\times$11 pixels.

We then selectively mask 50\% of the patches with the highest loss in each image \(I\). This selective masking process is repeated until all the images in this partition are masked.

\subsection{Models}
We chose a U-Net model with a ResNet34 backbone \citep{he2016deep} for our experiments. The U-Net is widely used in many generative models, most notably diffusion models \citep{rombach2022high, dhariwal2021diffusion, ho2020denoising}, and also excels in semantic segmentation tasks. We first initialize our U-Net with three output channels to predict the RGB values of the reconstructed image. Once the pretraining is complete, we initialize another downstream U-Net with the number of classes as the output channel count. The trained weights from the pretraining image reconstruction model are then copied to the downstream model except for the last layer to initialize the downstream training.

We use the Pytorch-toolbelt’s \citep{Khvedchenya_Eugene_2019_PyTorch_Toolbelt} implementation of the U-Net model and our ImageNet pretrained baseline uses their provided ImageNet pretrained ResNet34 weights.

\subsection{Datasets}
We chose four distinct datasets to test our selective masking method: Pascal VOC, Cityscapes, Nassar 2020 and Sugarbeets 2016. The images used for downstream training/testing are not excluded from the SSL pretraining since the image reconstruction pretraining process does not reveal the semantic segmentation labels to our model. Furthermore, using all the collected images to perform pretraining in a real-world workflow makes practical sense.

\textbf{Pascal VOC} is a general dataset with images commonly focusing on one or a few main objects. We use all 17,125 images in its \textsl{2012} subset for SSL pretraining and the 2,913 images with semantic segmentation labels for downstream training. We use the \textsl{2007} subset's \textsl{test} set to test the downstream performance.

\textbf{Cityscapes} is a street scene dataset with considerably higher resolution and object density than Pascal VOC. Cityscapes images are usually not focused on any particular object but instead contain many vehicles, signs and people to be detected. We use all images from the \textsl{leftImg8bit\_trainextra} subset (19,997 images) and the \textsl{leftImg8bit\_trainvaltest} subset (5,000 images) for SSL pretraining. The \textsl{train} set (2,975 images) of the \textsl{leftImg8bit\_trainextra} subset is used for downstream training while its \textsl{val} set (500 images) is used for downstream testing. The \textsl{test} set is not used in the downstream task as their labels are not publicly available.

\textbf{Nassar 2020} has a 29,395$\times$90,599 orthomosaic RGB whole-field image which is tiled into 40,710 non-overlapping 256$\times$256 tiles for SSL pretraining. This weed segmentation dataset is easier than the other above-mentioned datasets where a model with no pretraining can reach high accuracy. We choose only 1/5 of Nassar’s labelled \textsl{train/val} set (389 images) for downstream training to demonstrate the performance difference between pretraining methods on this small downstream dataset. All 266 tiles in the Nassar's \textsl{test} set are used in downstream testing.

\begin{figure}[h!]
\centering
\begin{minipage}{\textwidth}
	\includegraphics[width=0.08\textwidth]{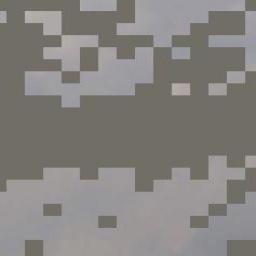}%
	\includegraphics[width=0.08\textwidth]{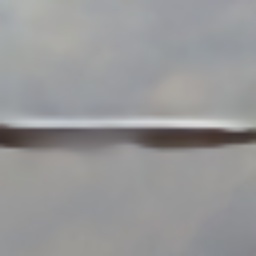}%
	\includegraphics[width=0.08\textwidth]{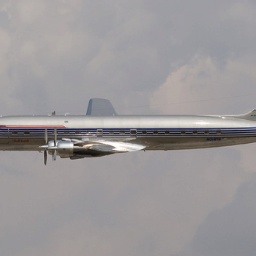}\hfil
	\includegraphics[width=0.08\textwidth]{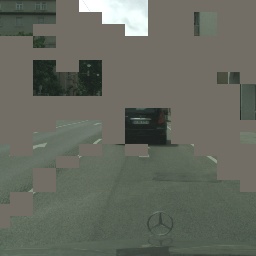}%
	\includegraphics[width=0.08\textwidth]{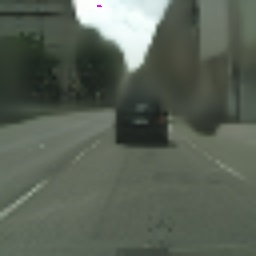}%
	\includegraphics[width=0.08\textwidth]{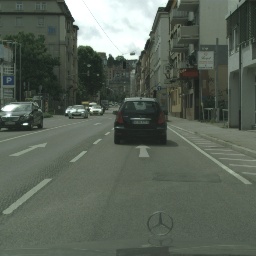}\hfil
	\includegraphics[width=0.08\textwidth]{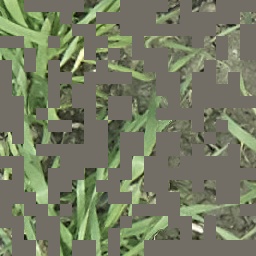}%
	\includegraphics[width=0.08\textwidth]{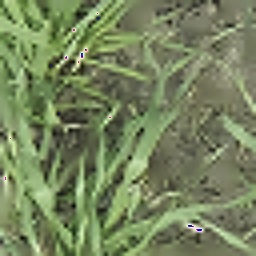}%
	\includegraphics[width=0.08\textwidth]{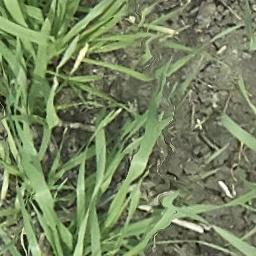}\hfil
	\includegraphics[width=0.08\textwidth]{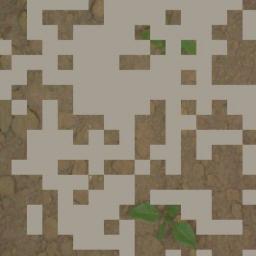}%
	\includegraphics[width=0.08\textwidth]{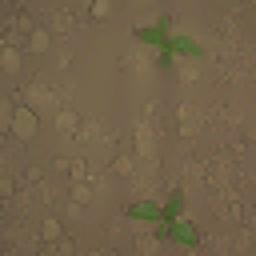}%
	\includegraphics[width=0.08\textwidth]{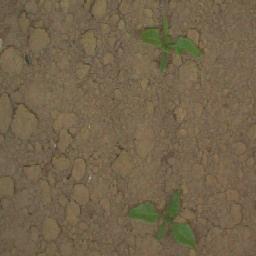}\\
	\includegraphics[width=0.08\textwidth]{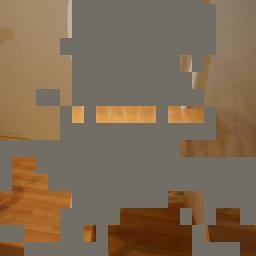}%
	\includegraphics[width=0.08\textwidth]{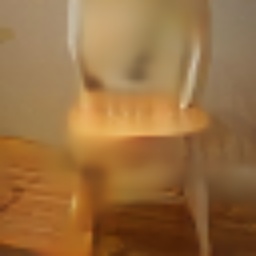}%
	\includegraphics[width=0.08\textwidth]{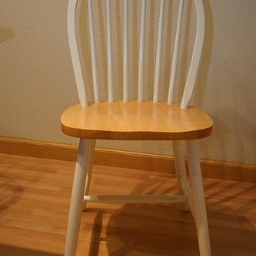}\hfil
	\includegraphics[width=0.08\textwidth]{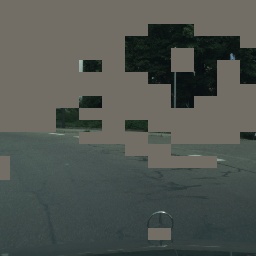}%
	\includegraphics[width=0.08\textwidth]{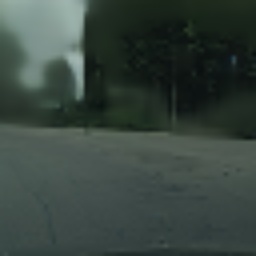}%
	\includegraphics[width=0.08\textwidth]{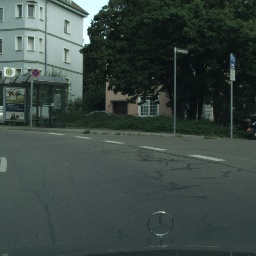}\hfil
	\includegraphics[width=0.08\textwidth]{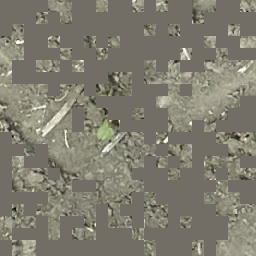}%
	\includegraphics[width=0.08\textwidth]{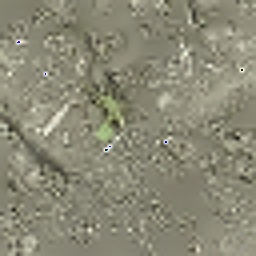}%
	\includegraphics[width=0.08\textwidth]{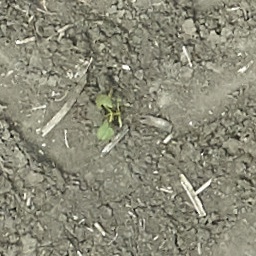}\hfil
	\includegraphics[width=0.08\textwidth]{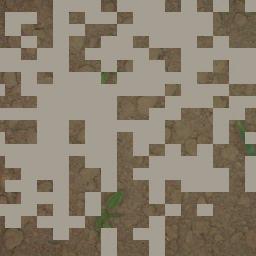}%
	\includegraphics[width=0.08\textwidth]{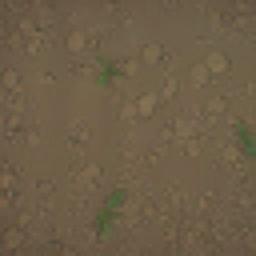}%
	\includegraphics[width=0.08\textwidth]{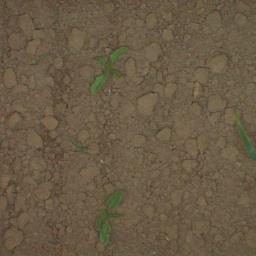}\\
	\includegraphics[width=0.08\textwidth]{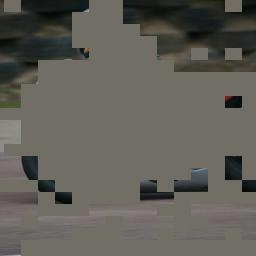}%
	\includegraphics[width=0.08\textwidth]{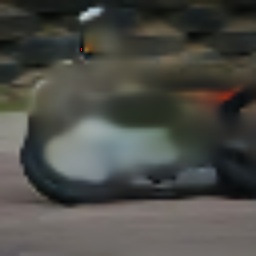}%
	\includegraphics[width=0.08\textwidth]{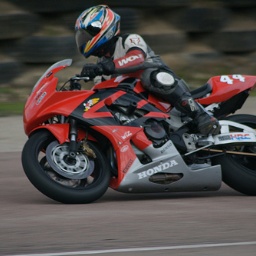}\hfil
	\includegraphics[width=0.08\textwidth]{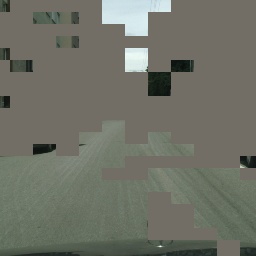}%
	\includegraphics[width=0.08\textwidth]{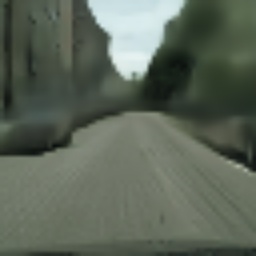}%
	\includegraphics[width=0.08\textwidth]{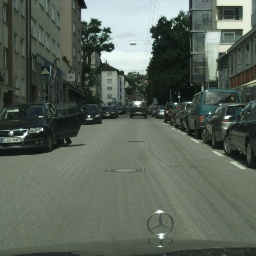}\hfil
	\includegraphics[width=0.08\textwidth]{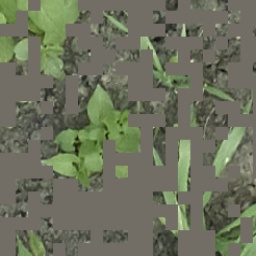}%
	\includegraphics[width=0.08\textwidth]{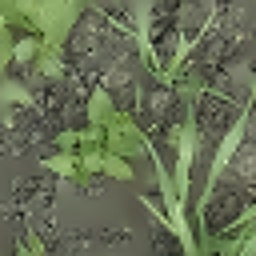}%
	\includegraphics[width=0.08\textwidth]{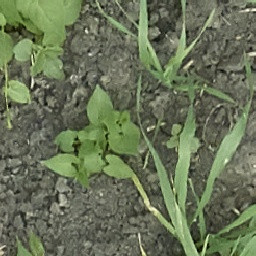}\hfil
	\includegraphics[width=0.08\textwidth]{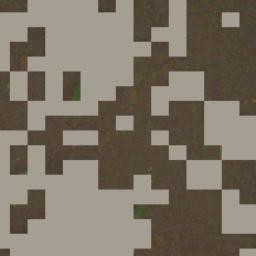}%
	\includegraphics[width=0.08\textwidth]{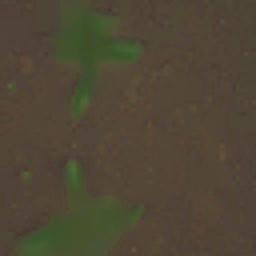}%
	\includegraphics[width=0.08\textwidth]{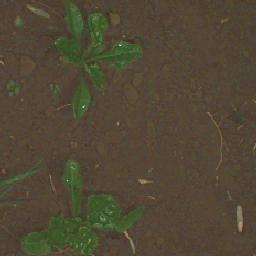}\\
	\includegraphics[width=0.08\textwidth]{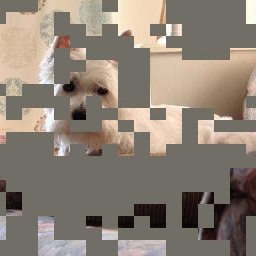}%
	\includegraphics[width=0.08\textwidth]{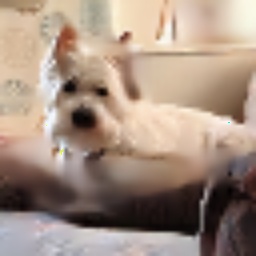}%
	\includegraphics[width=0.08\textwidth]{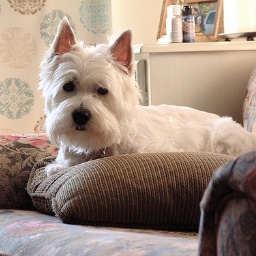}\hfil
	\includegraphics[width=0.08\textwidth]{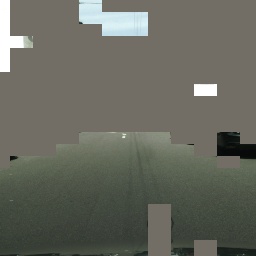}%
	\includegraphics[width=0.08\textwidth]{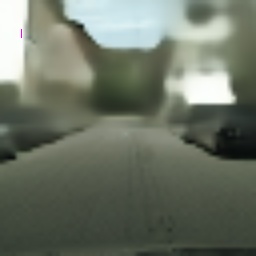}%
	\includegraphics[width=0.08\textwidth]{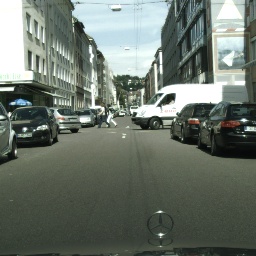}\hfil
	\includegraphics[width=0.08\textwidth]{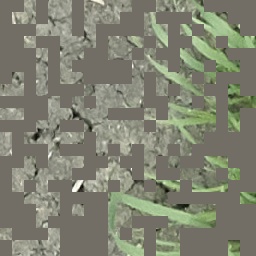}%
	\includegraphics[width=0.08\textwidth]{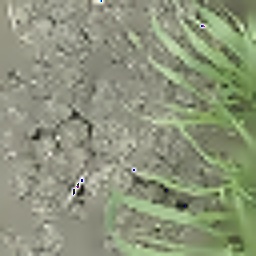}%
	\includegraphics[width=0.08\textwidth]{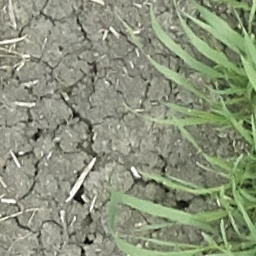}\hfil
	\includegraphics[width=0.08\textwidth]{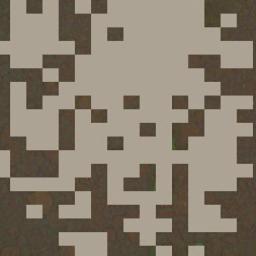}%
	\includegraphics[width=0.08\textwidth]{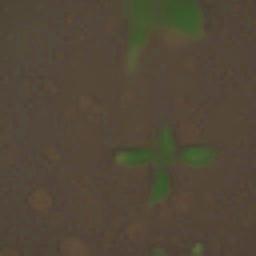}%
	\includegraphics[width=0.08\textwidth]{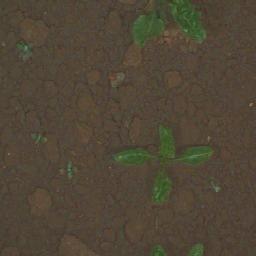}\\
	\centering
	\hfil(a)\hfil(b)\hfil(c)\hfil(d)\hfil
\end{minipage}

\caption[Selectively masked and reconstructed partition 9 examples from PascalVOC, Cityscapes, Nassar 2020, and Sugarbeets 2016.]{Selectively masked and reconstructed partition 9 examples from PascalVOC (a), Cityscapes (b), Nassar 2020 (c), and Sugarbeets 2016 (d). The brightness of Sugarbeets images is increased to improve visibility. The images in each column from left to right are: the selectively masked image, the reconstructed image, and the original image. Examples are randomly cropped to 256$\times$256 for validation, resulting in varied masked ratios. The Pascal VOC (a) and Cityscapes (b) examples show the selective masking process actively masks the entirety or majority of non-background object patches.}
\label{fig:rcon}
\end{figure}

\textbf{Sugarbeets 2016} is another weed segmentation dataset but is made challenging by poorly lit, low-contrast images containing sparsely distributed, small saplings making this dataset more challenging than Nassar. We use all 12,330 images for SSL pretraining and the 757 images that contain weed plants for the downstream task with a 7:1:2 train/val/test split. In Figure \ref{fig:comp_mask}, \ref{fig:rcon}, and \ref{fig:rcon_result}, we artificially increase the brightness of the example images for better visibility. The pretraining and downstream tasks use the unaltered original images.

\subsection{Experiments}

\subsubsection*{Image Reconstruction Pretraining}
We train the image reconstruction model on each partition for 1,000 epochs for selective masking pretraining. Since the size of each pretraining dataset is different, we set the batch size for each selective masking pretraining to be smaller than its partition size, which is 1/10 of the entire pretraining dataset. The random masking pretraining shares the number of epochs and batch sizes of their selective masking counterpart to ensure fair experiments. Table \ref{table:bs} lists the pretraining batch sizes for each dataset.

We use an MS-SSIM and L1 combo loss for the image reconstruction pretraining.
\begin{equation}
L = (1-MS{\text -}SSIM)+ L1
\end{equation}
The MS-SSIM loss helps the model to learn spatial features while the L1 loss helps to preserve color and luminance \citep{karnam2020self}. We use horizontal flip, color jitter and random crop augmentations in pretraining.

We compare our selective masking method with three baselines: random masking, supervised ImageNet pretraining and no pretraining. The experimented pretraining conditions are:

\begin{itemize}
	\item \textbf{Selective Masking Image Reconstruction}
	\item \textbf{Random Masking Image Reconstruction}
	\item \textbf{Supervised ImageNet Pretraining}
	\item \textbf{No Pretraining}
\end{itemize}

\subsubsection*{Downstream Task}
We use downstream segmentation mIoU and class IoU to evaluate the performance of different pretraining methods. Downstream training uses horizontal flip, color jitter, random crop augmentations and the Jaccard loss. Each downstream experiment trains for 500 epochs with a batch size of 8 and is repeated three times to minimize randomness using three randomly chosen seeds: 0, 1234, 3741. We report the average of three runs in the downstream results.

Our proposed selective masking image reconstruction method uses the same dataset for pretraining and the downstream task (self-pretraining). Even though collecting images in a semantic segmentation workflow is considerably cheaper than labelling images, the overall budget could be further decreased if we can perform pretraining on another similar dataset and only collect enough images for the downstream task (cross-pretraining). We test this for similar datasets only: pretraining on Cityscapes and downstream testing on Pascal, pretraining on Sugarbeets and downstream testing on Nassar, etc. The reported downstream results include:

\begin{itemize}
	\item \textbf{Downstream accuracy for the four proposed pretraining conditions}
	\item \textbf{Downstream accuracy for the lowest-performing classes}
	\item \textbf{Downstream accuracy for self-pretraining and cross-pretraining}
\end{itemize}

\section{Downstream Results}
\label{sec:results}

We show the downstream segmentation performance (mIoU) in Table \ref{table:main_result}. Our proposed Selective Masking Image Reconstruction pretraining method outperforms random masking image reconstruction pretraining and supervised ImageNet pretraining for all four tested datasets with clear margins. Selective masking shows a 4.5\% mIoU gain on Pascal VOC, a 2.9\% gain on Cityscapes and a 3.3\% gain on Nassar 2020 over supervised ImageNet pretraining. On the Sugarbeets 2016 dataset, selective masking beats random masking by 2.5\% while supervised ImageNet pretraining performed the worst (1.3\% lower than no pretraining).

Even though supervised ImageNet pretraining outperforms random masking on Pascal and Cityscapes, it shows no advantage over random masking on the weed segmentation dataset, Nassar. Another interesting observation is that the U-Net model performed poorly on Pascal and Cityscapes without pretraining, but the performance gap is considerably narrowed on the two weed segmentation datasets.

Table \ref{table:pascal_result} shows the six classes with the lowest IoU from the Pascal downstream experiments. Our selective masking method outperforms other baselines on all low-performing classes except for the potted plant class where it is 3.5\% lower than random masking. The horse class shows the highest performance boost where selective masking yields a 26.1\% gain over supervised ImageNet pretraining. Selective masking achieves a class IoU of 0.286 on the lowest-performing chair class which is 19.7\% higher than supervised ImageNet pretraining (0.239), the second highest. Overall, our selective masking method shows clear margins ($>$2.9\%) on 4 out of the 6 lowest-performing classes in Pascal VOC.

We compare the downstream performance of the six lowest-performing classes in the Cityscapes experiment in Table \ref{table:city_result}. Our selective masking method performed best in five classes except for the train class with the second highest performance (0.394) behind random masking (0.423). Selective masking shows a noticeable 9.4\% advantage for the motorcycle class which is the lowest-performing class for all other baselines. Out of six classes, selective masking shows an IoU gain of over 2.3\% for three.

We show the downstream foreground vegetation class IoU for the two weed segmentation datasets in Table \ref{table:nassar_result} (Nassar) and Table \ref{table:sugar_result} (Sugarbeets). Selective masking shows the highest IoU for the crop and the weed class for both datasets. On the more balanced Nassar dataset, selective masking yields around 5.5\% margins over random masking for both foreground classes. Selective masking outperforms random masking by 5.3\% for the crop class and 3.9\% for the underrepresented weed class on Sugarbeets. Similar to the overall result in Table \ref{table:main_result}, supervised ImageNet pretraining performs comparably to random masking on Nassar while worse than all other methods on Sugarbeets.

We compare the downstream performance of selective masking self-pretraining and cross-pretraining in Table \ref{table:cross_result}. The results show that self-pretraining yields better performance for all four datasets, with margins over 2.4\%.

\section{Discussion}

The results in Section \ref{sec:results} show our proposed Selective Masking Image Reconstruction pretraining method consistently outperforms traditional random masking image reconstruction and supervised ImageNet pretraining with substantial margins. Our method also consistently boosts the performance of the lowest-performing class in most cases. The most underrepresented class, like the motorcycle class in Cityscapes, tends to have the lowest accuracy due to the lack of positive samples for the model to learn. Our proposed selective masking image reconstruction pretraining method provides a low-cost way to improve accuracy on underrepresented classes without creating synthetic data or using complex dataset/loss balancing tricks.

Table \ref{table:main_result}, \ref{table:pascal_result}, \ref{table:city_result} show that the traditional random masking image reconstruction method is not effective on complex general datasets like Pascal VOC and Cityscapes when pretrained on smaller datasets. The SSL pretraining generally has a different task from the downstream training, making the self-supervised pretraining process far less efficient than fully-supervised training. Combining a less efficient training process with a less efficient random masking scheme makes traditional SSL processes extremely data-hungry. As mentioned in Section \ref{sec:intro}, a high-capacity model has high generalizability but is expensive to train, memory intensive and leads to slower inference. Many computer vision tasks like weed detection benefit little from model generalizability, and others like autonomous driving value fast inference and low training and operation costs. Our proposed selective masking image reconstruction method provides an efficient and low-cost way to improve the downstream performance of CNN models.

The weed segmentation experiments shown in Table \ref{table:main_result}, \ref{table:nassar_result}, \ref{table:sugar_result} show that supervised ImageNet pretraining does not transfer well to weed segmentation. Many domain-specific tasks suffer from this problem as ImageNet contains little to no domain-specific information. Our cross-pretraining experiment (Table \ref{table:cross_result}) shows that self-pretraining performs better and should be preferred when possible. However, the selective masking cross-pretraining experiments show comparable or better performance than supervised ImageNet pretraining. Selective masking cross-pretraining could be a low-cost alternative to improve downstream performance for domain-specific tasks when in-domain pretraining images are scarce.

\subsection{Reconstruction Results}
\begin{figure}[h!]
\centering
\begin{minipage}{\linewidth}
	\includegraphics[width=0.18\linewidth]{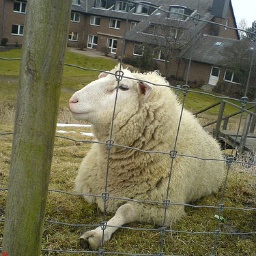}\hfil
	\includegraphics[width=0.18\linewidth]{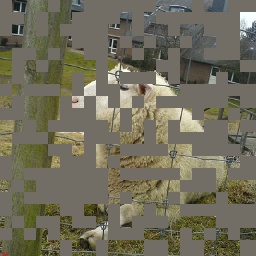}%
	\includegraphics[width=0.18\linewidth]{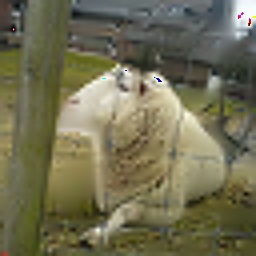}\hfil
	\includegraphics[width=0.18\linewidth]{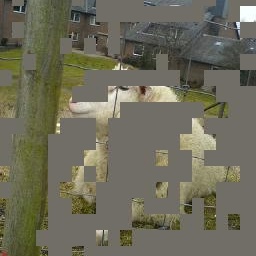}%
	\includegraphics[width=0.18\linewidth]{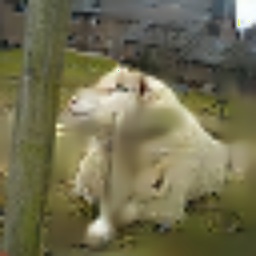} \\
	\includegraphics[width=0.18\linewidth]{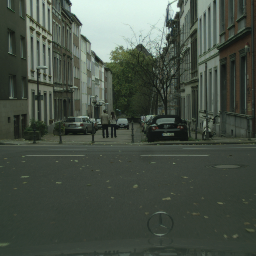}\hfil
	\includegraphics[width=0.18\linewidth]{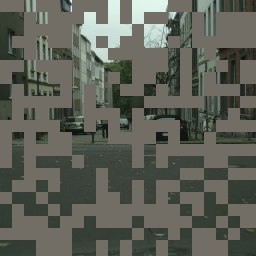}%
	\includegraphics[width=0.18\linewidth]{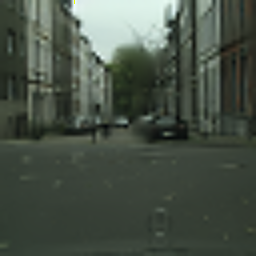}\hfil
	\includegraphics[width=0.18\linewidth]{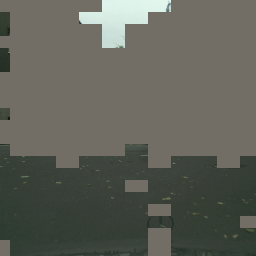}%
	\includegraphics[width=0.18\linewidth]{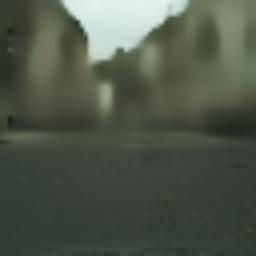} \\
	\includegraphics[width=0.18\linewidth]{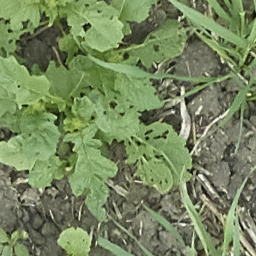}\hfil
	\includegraphics[width=0.18\linewidth]{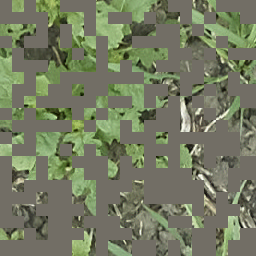}%
	\includegraphics[width=0.18\linewidth]{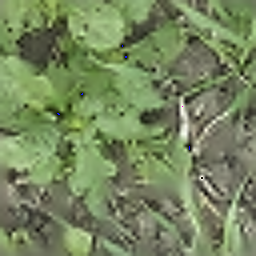}\hfil
	\includegraphics[width=0.18\linewidth]{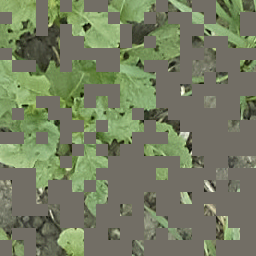}%
	\includegraphics[width=0.18\linewidth]{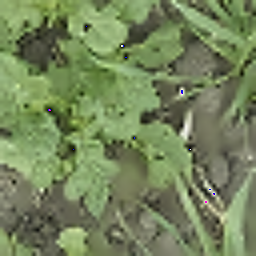} \\
	\includegraphics[width=0.18\linewidth]{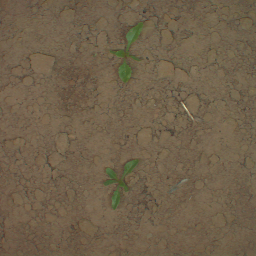}\hfil
	\includegraphics[width=0.18\linewidth]{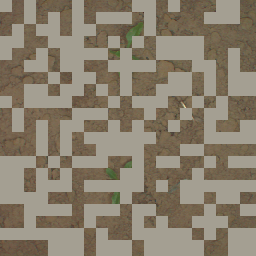}%
	\includegraphics[width=0.18\linewidth]{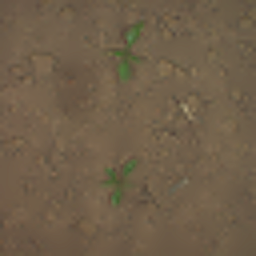}\hfil
	\includegraphics[width=0.18\linewidth]{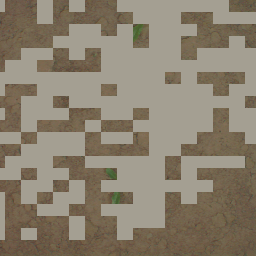}%
	\includegraphics[width=0.18\linewidth]{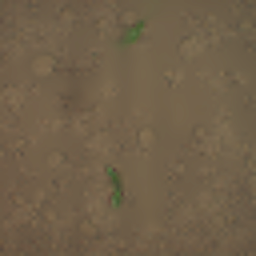} \\
	\hfil$~~~~~~~$(a)\hfil(b)\hfil$~~~~~~~~~$(c)\hfil
\end{minipage}

\caption[Comparison of selective masking and random masking reconstruction examples from partition 9.]{Comparison of selective masking and random masking reconstruction examples from partition 9 showing the original image (a), randomly masked and reconstructed images (b), and selectively masked and reconstructed images (c). The examples from the top to the bottom row are from Pascal VOC, Cityscapes, Nassar 2020, Sugarbeets 2016. The brightness of Sugarbeets images is increased to improve visibility.}
\label{fig:rcon_result}
\end{figure}

Table \ref{table:recon_result} compares the validation reconstruction loss between selective masking and random masking for partition 9. Selective masking results in considerably higher reconstruction loss for Pascal and Cityscape compared to random masking. This can be explained by the Cityscapes (second row) examples shown in Figure \ref{fig:rcon_result} (c) where selective masking is significantly more condensed, covering the entirety of the two buildings. Interestingly, we can observe from the reconstructed image that without surrounding patches to provide a hint about the masked regions, the model still predicted two buildings but the predictions are very different from the original buildings. This reconstruction sample shows the model successfully learnt visual features of buildings but have lower patch SSIM values since the reconstructed buildings are different from the original ones. This contrasts with the random masking sample shown in Figure \ref{fig:rcon_result} (b), where the model could simply learn to connect the dots between the evenly distributed masked patches to lower reconstruction loss, resulting in the model learning fewer features. Weed segmentation datasets show similar trends but are less noticeable, especially for images with high vegetation coverage, such as the Nassar example (third row) shown in Figure \ref{fig:rcon_result}.

The reconstruction loss comparison shows two things. First, proper pretraining task difficulty is important: a more difficult task forces the model to learn useful features instead of exploiting shortcuts in the task’s design. Secondly, pretraining performance does not directly correlate to downstream performance.

\subsection{Related Works}
\label{sec:related}

\subsection*{Self-supervised Learning}
An SSL process usually consists of a pretraining task and a downstream task. The model is trained on unlabelled data to learn representations in the pretraining task. The pretrained model weights are then used to initialize the model for the downstream task, which is usually trained fully-supervised on a labelled dataset.

Popular approaches to self-supervised learning for computer vision include contrastive methods like Simple Siamese Representation Learning (SimSiam) \citep{chen2021exploring}, as well as generative methods like Generative Adversarial Networks (GANs) \citep{goodfellow2020generative}, Masked Autoencoders (MAE) \citep{he2022masked} and Bidirectional Encoder (BEiT) \citep{bao2021beit}. GANs and many contrastive methods are very promising but can be computationally intensive due to the presence of multiple networks or the heavy reliance on large batch sizes and augmentations \citep{chen2020simple, chen2021exploring, grill2020bootstrap}.

\subsection*{Masked Image Modeling}
Masked image modelling (MIM) SSL methods train the model to perform image reconstruction on masked images \citep{liu2021self} to learn representations from these unlabelled images. Most MIM research, like MAE \citep{he2022masked} and BEiT \citep{bao2021beit}, was inspired by LLM pretraining and is designed only for transformer models. Some works attempt to merge autoencoder-based methods with CNN architectures \citep{woo2023convnext, tian2023designing, gao2022convmae}. Karnam \etal \citep{karnam2020self} proposed directly using a U-Net \citep{ronneberger2015u} to reconstruct images and reported improved performance on the downstream semantic segmentation task. Chamberlain \etal \citep{chamberlain2023self} extended this research by studying masking ratios and various masking methods.

\subsection*{Data Augmentation in SSL}
Due to the lack of labels, most SSL methods rely heavily on data augmentations to provide information for the model to learn. Contrastive models using the Siamese Network architecture commonly use various augmentation methods to generate different views of the same image \citep{chen2021exploring}. Some SSL methods are especially sensitive to certain augmentations: SimCLR \citep{chen2020simple} used random crop/resize, color jittering and Gaussian blur and identified color distortion as the key component to success. BYOL \citep{grill2020bootstrap} has a more comprehensive augmentation pipeline, including horizontal flip, color dropping, and solarization, in addition to the ones found in SimCLR. Some methods also include grayscale conversion in the augmentation pipeline, such as Barlow Twins \citep{zbontar2021barlow}.

Generative MIM methods simply augment input images by masking/corruption. He \etal \citep{he2022masked} discussed three masking strategies: random sampling, block-wise sampling and grid-wise sampling. Block-wise sampling masks large random blocks, while grid-wise sampling drops uniformly distributed grid patches. The author found random sampling the most effective due to its ability to mask larger portions of the image. Chamberlain \etal \citep{chamberlain2023self} discussed dropping random color channels instead of masking whole pixels. Xie \etal \citep{xie2024rethinking} proposed leveraging radiological reports to perform selective masking in medical imaging, outperforming random masking MIM. We identified a gap in existing MIM studies where the potential of using model knowledge to perform selective masking remains mostly unexplored.

\subsection{Limitations and Future Works}
The performance gain from our Selective Masking Image Reconstruction method comes with added computational cost. The time spent on iterative model training remains the same as random masking since we only train on each partition once. Inferencing on five randomly masked samples and then sorting the resulting average patch loss map contribute to the added computation cost. The incremental cost is marginal for our intended low-budget pretraining scenarios due to low image and low patch-per-image counts, but may become considerable for very large datasets.

Active learning (AL) is another tool to improve deep learning data efficiency by minimizing the labelling cost for the supervised downstream task. Uncertainty sampling active learning methods are generally iterative processes where the model is initialized with a very small randomly chosen labelled dataset. The quality of the initialization training is crucial as it influences the quality of the uncertainty measurement used in data ranking but is often limited by the very small initialization dataset. We could use self-supervised pretraining to improve the performance of the initialization training, leading to faster convergence and better accuracy. Furthermore, regional active learning methods designed for semantic segmentation typically generate pixel/patch-wise uncertainty maps to help select the labelling regions. By combining the uncertainty map with our reconstruction patch loss map, we could diversify the uncertainty calculation. This AL/SSL hybrid method could potentially improve the quality of the region/image selection process.

\subsection{Conclusion}
To conclude, our proposed Selective Masking Image Reconstruction method provides an effective and practical solution to improve end-to-end semantic segmentation workflows, especially for scenarios that require limited model capacity to meet inference speed and computational resource requirements. Our proposed pretraining method could also be helpful for specialized use-cases where general large-scale pretraining does not work well, due to the lack of domain-specific information in the pretraining dataset.

\section*{Conflict of Interest Statement}

The authors declare that the research was conducted in the absence of any commercial or financial relationships that could be construed as a potential conflict of interest.

\section*{Author Contributions}

YW: Writing - original draft, Conceptualization, Formal analysis, Investigation, Methodology, Software, Visualization; IS: Writing - review \& editing, Funding acquisition, Project administration, Resources, Supervision

\section*{Funding}
This research was undertaken thanks in part to funding from the Canada First Research Excellence Fund.

\section*{Acknowledgments}
Thank you to the staff at the Kernen Crop Research Farm for the field trials utilized in this study.

\section*{Data Availability Statement}

The Pascal VOC dataset can be found at \url{https://web.archive.org/web/20140815141459/http://pascallin.ecs.soton.ac.uk/challenges/VOC/voc2012/index.html}.

The Cityscapes dataset can be found at \url{https://github.com/mcordts/cityscapesScripts}.

The Nassar 2020 dataset can be found at \url{https://www.kaggle.com/datasets/yueminwang/nassar-2020}.

The Sugarbeets 2016 dataset can be found at \url{https://www.ipb.uni-bonn.de/data/sugarbeets2016/index.html}.

\bibliographystyle{unsrtnat}
\bibliography{main}  






\section*{Tables}

\begin{table}[h!]
\centering
\begin{tabular}{c c c c}

Pascal & Cityscapes & Nassar & Sugarbeets \\
\Xhline{2\arrayrulewidth}
128 & 128 & 256 & 64\\
\end{tabular}
\caption{Batch sizes are chosen to be smaller than the partition size of the corresponding dataset. The partition size is 1/10 of the pretraining dataset.}
\label{table:bs}
\end{table}

\begin{table}[h!]
\centering
\begin{tabular}{c c c c c}

Dataset & Selective & Random & ImageNet & None\\
\Xhline{2\arrayrulewidth}
Pascal & \textbf{0.603} & 0.566  & 0.577 & 0.4111\\

Cityscapes & \textbf{0.608} & 0.572  & 0.591 & 0.541\\

Nassar & \textbf{0.847} & 0.819  & 0.820 & 0.806\\

Sugarbeets & \textbf{0.739} & 0.721  & 0.689 & 0.698\\
\end{tabular}
\caption{Downstream semantic segmentation results (mIoU). We compare \textbf{Selective} masking image reconstruction, \textbf{Random} masking image reconstruction, \textbf{ImageNet} supervised pretraining and \textbf{None} as no pretraining. We report the average of three runs.}
\label{table:main_result}
\end{table}

\begin{table}[h!]
\centering
\begin{tabular}{c c c c c}

Class & Selective & Random & ImageNet & None\\
\Xhline{2\arrayrulewidth}
dog & \textbf{0.329} & 0.293 & 0.328 & 0.192 \\
horse & \textbf{0.502} & 0.395 & 0.398 & 0.294 \\
potted plant & 0.332 & \textbf{0.344} & 0.252 & 0.073 \\
sofa & \textbf{0.430} & 0.322 & 0.418 & 0.212 \\
bottle & \textbf{0.403} & 0.386 & 0.282 & 0.073 \\
chair & \textbf{0.286} & 0.188 & 0.239 & 0.116 \\
\end{tabular}
\caption{Pascal VOC downstream results (IoU) for six classes with the lowest class IoU.}
\label{table:pascal_result}
\end{table}

\begin{table}[h!]
\centering
\begin{tabular}{c c c c c}

Class & Selective & Random & ImageNet & None \\
\Xhline{2\arrayrulewidth}
rider & \textbf{0.491} & 0.411 & 0.480 & 0.357 \\
train & 0.394 & \textbf{0.423} & 0.344 & 0.382 \\
motorcycle & \textbf{0.360} & 0.288 & 0.329 & 0.259 \\
fence & \textbf{0.336} & 0.287 & 0.334 & 0.262 \\
pole & \textbf{0.416} & 0.382 & 0.397 & 0.344 \\
traffic light & \textbf{0.476} & 0.408 & 0.467 & 0.396 \\
\end{tabular}
\caption{Cityscapes downstream results (IoU) for six classes with the lowest class IoU.}
\label{table:city_result} 
\end{table}

\begin{table}[h!]
\centering
\begin{tabular}{c c c c c}

Class & Selective & Random & ImageNet & None \\
\Xhline{2\arrayrulewidth}
crop & \textbf{0.833} & 0.792 & 0.797 & 0.784 \\
weed & \textbf{0.727} & 0.689 & 0.688 & 0.658 \\
\end{tabular}
\caption{Nassar downstream results (IoU) for classes crop and weed.}
\label{table:nassar_result}
\end{table}

\begin{table}[h!]
\centering
\begin{tabular}{c c c c c}

Class & Selective & Random & ImageNet & None \\
\Xhline{2\arrayrulewidth}
crop & \textbf{0.759} & 0.721 & 0.670 & 0.682 \\
weed & \textbf{0.476} & 0.458 & 0.418 & 0.432 \\
\end{tabular}
\caption{Sugarbeets downstream results (IoU) for classes crop and weed.}
\label{table:sugar_result}
\end{table}

\begin{table}[h!]
\centering
\begin{tabular}{c c c c}

Downstream & Self-pretrain & Cross-pretrain & ImageNet\\
\Xhline{2\arrayrulewidth}
Pascal & \textbf{0.603} & 0.589 & 0.577\\
Cityscapes & \textbf{0.608} & 0.592 & 0.591 \\
Nassar & \textbf{0.847} & 0.827 & 0.820\\
Sugarbeets & \textbf{0.739} & 0.718 & 0.689\\
\end{tabular}
\caption{Downstream semantic segmentation results (mIoU) of selective masking \textbf{Self-pretrain}ing, \textbf{Cross-pretrain}ing and supervised \textbf{ImageNet} pretraining. Self-pretraining uses the same pretraining (unlabelled) and downstream dataset while pretraining and downstream datasets are different for cross-pretraining}
\label{table:cross_result}
\end{table}

\begin{table}[h!]
\centering
\begin{tabular}{c c c}

Dataset & Selective & Random \\
\Xhline{2\arrayrulewidth}
Pascal & 0.613 & 0.267 \\

Cityscapes & 0.536 & 0.193 \\

Nassar & 0.526 & 0.433 \\

Sugarbeets & 0.345 & 0.246 \\
\end{tabular}
\caption{Validation reconstruction loss for partition 9.}
\label{table:recon_result}
\end{table}

\clearpage

\section*{Figure captions}


\end{document}